%% file: main.tex
\documentclass[runningheads]{llncs}
\usepackage[T1]{fontenc}
\usepackage{graphicx}
\usepackage{booktabs}
\usepackage{amssymb}
\usepackage{amsmath}
\usepackage{caption}
\usepackage{multirow}
\usepackage{threeparttable}
\usepackage{graphicx}
\usepackage{float}
\usepackage{subfig}
\usepackage{soul, color, xcolor}
\usepackage[misc]{ifsym}
\newcommand{\corr}{(\Letter)}

\usepackage{mwe}
\usepackage{hyperref}

\toctitle{C$^3$DE: Causal-Aware Collaborative Neural Controlled Differential Equation for Long-Term Urban Crowd Flow Prediction}
\tocauthor{Yuting Liu, Qiang Zhou, Hanzhe Li, Chenqi Gong, Jingjing Gu}

\begin{document}

\title{C$^3$DE: Causal-Aware Collaborative Neural Controlled Differential Equation for Long-Term Urban Crowd Flow Prediction}

\titlerunning{Causal-Aware Collaborative NCDE for Long-Term Crowd Flow Prediction}

\author{Yuting Liu\inst{1} \and Qiang Zhou\inst{1} \corr \and 
Hanzhe Li\inst{1} \and Chenqi Gong\inst{2} \and Jingjing Gu\inst{1}}

\authorrunning{Y. Liu et al.}

\institute{Nanjing University of Aeronautics and Astronautics, Nanjing, China \email{\{yuting\_liu,zhouqnuaacs,lihanzhe,gujingjing\}@nuaa.edu.cn}
\and
Chongqing University, Chongqing, China \\ \email{gcq@stu.cqu.edu.cn}}

\maketitle              

\input{01.Abstract}
\input{02.Introduction}

\input{03.Relatedwork}
\input{04.Preliminaries}
\input{05.Methodology}
\input{06.Experiments}
\input{07.Conclusion}

%
%
%
\bibliographystyle{splncs04}
%

\end{document}

%% file: 01.Abstract.tex
\begin{abstract}
Long-term urban crowd flow prediction suffers significantly from cumulative sampling errors, due to increased sequence lengths and sampling intervals, which inspired us to leverage Neural Controlled Differential Equations (NCDEs) to mitigate this issue. However, regarding the crucial influence of Points of Interest (POIs) evolution on long-term crowd flow, the multi-timescale asynchronous dynamics between crowd flow and POI distribution, coupled with latent spurious causality, poses challenges to applying NCDEs for long-term urban crowd flow prediction.
To this end, we propose \underline{C}ausal-aware \underline{C}ollaborative neural \underline{CDE} (C$^3$DE) to model the long-term dynamic of crowd flow.
Specifically, we introduce a dual-path NCDE as the backbone to effectively capture the asynchronous evolution of collaborative signals across multiple time scales.
Then, we design a dynamic correction mechanism with the counterfactual-based causal effect estimator to quantify the causal impact of POIs on crowd flow and minimize the accumulation of spurious correlations.
Finally, we leverage a predictor for long-term prediction with the fused collaborative signals of POI and crowd flow.
Extensive experiments on three real-world datasets demonstrate the superior performance of C$^3$DE, particularly in cities with notable flow fluctuations.

\keywords{Long-Term Urban Crowd Flow Prediction  \and Neural Controlled Differential Equation \and Counterfactual Inference.}
\end{abstract}

%% file: 02.Introduction.tex
\section{Introduction}
Urban development is a dynamic process driven by population growth, economic activities, and infrastructure development. 
As a key indicator of urban operations, urban crowd flow exhibits a significant continuous evolution trend. 
Exploring its long-term evolution helps reveal urban operating  patterns and provides valuable insights for traffic management and sustainable urban development.

Existing researches on long-term prediction~\cite{jiang2023pdformer,yu2023dsformer,zheng2023prediction} typically employed coarse-grained data with hourly or even longer intervals, in contrast to the high-frequency, minute-level data commonly used. 
Such coarse-grained data obscures the important urban dynamics and trends, leading to information loss and making it harder for models to capture crowd flow dynamics, resulting to suboptimal prediction performance. 
Consequently, we introduce a continuous modeling approach to better capture urban dynamics from coarse-grained data, enhancing prediction stability and accuracy.
Intuitively, it is essential to consider the evolution of urban structure in urban crowd flow prediction, which is mainly reflected in the changes of POI distribution~\cite{li2022user,zeng2022causal}. 
For example, the construction of a new commercial center may attract higher pedestrian flow, while the renovation of an old residential area may affect the surrounding traffic flow. 

However, modeling the impact of POI distribution on crowd flow (also referred to as collaborative signals) from a continuous-time perspective poses the following two challenges:
\textbf{\textit{i) The multi-timescale asynchronous dynamics of collaborative signals increase the difficulty of modeling spatio-temporal dependencies in urban dynamic systems.}}
The evolution of collaborative signals occurs across different time scales, with their dynamic changes unsynchronized.
Specifically, changes in low-frequency POI distributions gradually manifest in the high-frequency crowd flow patterns.
This cross-scale influence is often reflected in significant crowd flow variations across multiple timestamps, which significantly increases the difficulty of modeling the multi-scale asynchronous dynamic and revealing dynamic patterns in a urban system.
\textbf{\textit{ii) The accumulation of spurious correlations between collaborative signals complicates the identification of true causal relationships in continuous modeling.}} 
POI distribution and urban crowd flow often exhibit statistically spurious correlations, which may mislead the model.
From a discrete-time perspective, spurious correlations can be easily identified and removed through statistical methods like calculating correlation coefficients or Granger causality tests~\cite{diks2006new}.
However, in continuous-time modeling, where time is treated as a continuous variable and dynamics are learned through differential equations\cite{chen2018neural}, spurious correlations can be amplified during long-term integration.
Moreover, minor perturbations in continuous time can significantly impact the overall system, further complicating the accurate identification of true causality.

To this end, we propose a \textbf{C}ausal-aware \textbf{C}ollaborative Neural \textbf{C}ontrolled \textbf{D}ifferential \textbf{E}quations framework (C$^3$DE) for long-term urban crowd flow prediction. 
Specifically, we propose a collaborative neural controlled differential equation (NCDE) with a dual-path architecture to capture the dynamic evolution of collaborative signals across different timescales in continuous time. 
With the continuous-time integration property of NCDE, the asynchronous dynamics of collaborative signals can be effectively modeled.
Furthermore, we design a counterfactual-based causal effect estimator to simulate urban dynamics under different POI distribution interventions, enabling a quantitative assessment of each POI category’s direct impact on crowd flow.
To mitigate the accumulation of spurious correlations among collaborative signals, we incorporate causal effect values into the NCDE framework and introduce a causal effect-based dynamic correction mechanism. 
By computing causal influences across multiple time steps and feeding them back into POI representations, the mechanism effectively minimizes interference from spurious POIs, alleviates the amplification of spurious correlations, and enhances the model’s robustness and reliability in the long-term prediction task. 

Overall, our contributions can be summarized as follows:
\begin{itemize}
    \item[-] To the best of our knowledge, C$^3$DE, is the first to simulate the evolution of collaborative signals and explore the underlying causal mechanisms for long-term urban crowd flow prediction.
    \item[-] We propose a collaborative NCDE with a dual-path architecture to effectively capture the asynchronous evolution of collaborative signals across multiple timescales. 
    \item[-] We design a counterfactual-based causal effect estimator to quantify the causal impact of POIs on crowd flow and introduce a causal effect-based dynamic correction mechanism to reduce the accumulation of spurious correlations.
    \item[-] Extensive experiments on three real-world datasets demonstrate that C$^3$DE offers a significant advantage in modeling crowd flow dynamics, particularly in cities with notable flow fluctuations.
\end{itemize}

%% file: 03.Relatedwork.tex
\section{Related Work}
\subsubsection{Urban Crowd Flow Prediction.} 
Recently, urban crowd flow prediction~\cite{chen2024multivariate,liu2024frequency} has become a critical research topic, relying on historical flow data and using Gated Recurrent Unit (GRU) and Graph Neural Networks (GNNs) to learn spatio-temporal features.
Traditional spatio-temporal GNNs, like STGCN~\cite{yu2018spatio} and STSGCN~\cite{song2020spatial}, used predefined graph structures to capture spatial dependencies but often fail to capture the hidden ones.
To address this, methods based on adaptive graph structures introduced learnable adaptive adjacency matrices, enabling capturing the dynamics of node relationships~\cite{bai2020adaptive,shao2022decoupled,wu2019graph}.
In addition, considering that urban structure, i.e., POI distribution, significantly affect crowd mobility patterns, some works have incorporated it into flow pattern modeling~\cite{liang2018geoman,rong2021inferring}.
For example, GeoMAN~\cite{liang2018geoman} treated POI as a spatial feature to capture spatial correlations within regions. 
GSTE-DF~\cite{rong2021inferring} utilizes POI data to uncover differences and similarities between regions for inferring origin-destination flows.
Although these works achieved some success, they treated POI as a static feature and ignored the dynamics of POI distribution in cities.

\vspace{-.5cm}

\subsubsection{Neural Ordinary Differential Equations.} 
\cite{chen2018neural} first combined neural networks with Ordinary Differential Equations (ODE) and proposed Neural ODEs to model continuous dynamics, which has been widely used in the fields of time series prediction~\cite{gravina2024temporal,jin2022multivariate}, continuous dynamic systems~\cite{huang2020learning,huang2021coupled}.
~\cite{fang2021spatial} proposed tensor-based ODEs to capture spatio-temporal dynamics, overcoming the limitations of graph convolutions in modeling long-range spatial dependencies and semantic connections. 
~\cite{choi2022graph} designed two types of Neural Controlled Differential Equations to handle temporal and spatial dependencies separately. 
Additionally,~\cite{long2024unveiling} proposed STDDE, which incorporates delayed states into NCDE, allowing it to model time delays in spatio-temporal information propagation.

\vspace{-.5cm}

\subsubsection{Counterfactual Inference.}
The main goal of counterfactual inference is to analyze potential outcomes through hypothetical interventions and answer "What would have happened if the situation had been different?"~\cite{chen2021human,tian2022debiasing}.
For example,~\cite{li2024beyond} proposed a counterfactual data augmentation-based causal explanation framework that identifies the true causal factors by constructing counterfactual data.~\cite{shao2023cube} introduced a counterfactual explanation method based on causal intervention, using a causal director to capture causal relationships in the distribution and guide counterfactual generation.
In this paper, We address the spurious correlations between collaborative signals from a counterfactual perspective.

%% file: 04.Preliminaries.tex
\section{Preliminary}

\subsection{Definitions and problem statement}
\textbf{Definition 1 (Urban Network).} The urban network is represented as a directed graph $G=(V, X, A)$, where $V=\left \{ 
 V_1, V_2, ..., V_N \right \} $ denotes $N$ regions in the city. 
 $X \in \mathbb{R}^{t^\prime \times N \times C} $ denotes the urban crowd flow across $N$ regions at $t^\prime$ time steps, where $t^\prime$ is measured in days and $C$ capturing hourly features.
 $A \in \mathbb{R}^{N\times N} $ is the adjacency matrix, which encodes the relationships between regions.
 
 \noindent \textbf{Definition 2 (POI Distribution).} The POI distribution is denoted as $P \in \mathbb{R}^{t^{\prime\prime} \times N \times K} $, where $t^{\prime\prime}$ is the time steps, measured in months. $K$ represents the number of POI categories, such as restaurants, shops and public facilities.

 \noindent \textbf{Problem Statement (Long-Term Urban Crowd Flow prediction).}
 Given the crowd flow for the past $T$ time steps and POI distribution for the past $M$ time steps, our goal is to learn a map function $\mathcal{F}(\cdot)$
 that capture the causal evolutionary relationship and predict the urban crowd flow for the next $S$ time steps. It can be formulated as follows:
 \begin{equation}
     \mathcal{F}^\ast = arg \min_{\mathcal{F}} \sum_{S} \ell (\mathcal{F}(X_{t^\prime-T+1:t^\prime}, P_{t^{\prime\prime}-M+1:t^{\prime\prime}}),X_{t^\prime+1:t^\prime+S}),
 \end{equation}
where $\mathcal{F}^\ast$ denote the function with the learned optimal parameters, and $\ell(\cdot)$ is the loss function.

In this work, we divide the map function $\mathcal{F}(\cdot)$ into two stages, i.e., a representation part $F(\cdot)$ to model the collaborative causal evolutionary relationship and a predictor $G(\cdot)$ to predict the future crowd flow.

 \subsection{Neural Differential Equation}
 \textbf{Neural ODEs.}
 Neural ODEs~\cite{chen2018neural} extend residual networks into the continuous time domain.
Given the input $X$, neural ODEs define a hidden state $h(t)$ that evolves over time $t$, as described by the following Riemann integral:
\begin{equation} \label{NODE}
    h(t)=h(0)+\int_{0}^{t}\frac{\mathrm{d} h}{\mathrm{d} t}  \mathrm{d}t=h(0)+\int_{0}^{t}f(h(t),t;\theta) \mathrm{d}t ,
\end{equation}
where a neural network $f(\cdot)$ with parameter $\theta$ parameterize the derivative of the hidden state, i.e., $\frac{\mathrm{d} h}{\mathrm{d} t}:=f(h(t),t;\theta) $.
The evolution process is computed using ODE solvers, such as the Euler method and Runge-Kutta. To improve efficiency, the adjoint sensitivity method is often employed to compute the parameter gradients via the adjoint equations, rather than direct backpropagation.

 \noindent \textbf{Neural CDEs.}
 Neural CDEs~\cite{kidger2020neural} are the extension of neural ODEs. Neural CDEs introduces an external control signal $X_t$, which drives the evolution of the hidden state $h(t)$, making it dependent on both its own dynamics and the control signal. Specifically, it can be expressed as:
 \begin{equation}
     h(t)=h(0)+\int_{0}^{t} f(h(t),t;\theta) \mathrm{d}X_t ,
 \end{equation}
 where $X_t$ is a continuous path defined in a Banach space, representing the external control signal. Different from Eq. \ref{NODE}, it represents a Riemann$-$Stieltjes integral, allowing to model the influence of control signal on system's evolution.

%% file: 05.Methodology.tex
\section{Methodology}
\begin{figure*}[t]
\centering  
\includegraphics[width=0.9\linewidth]{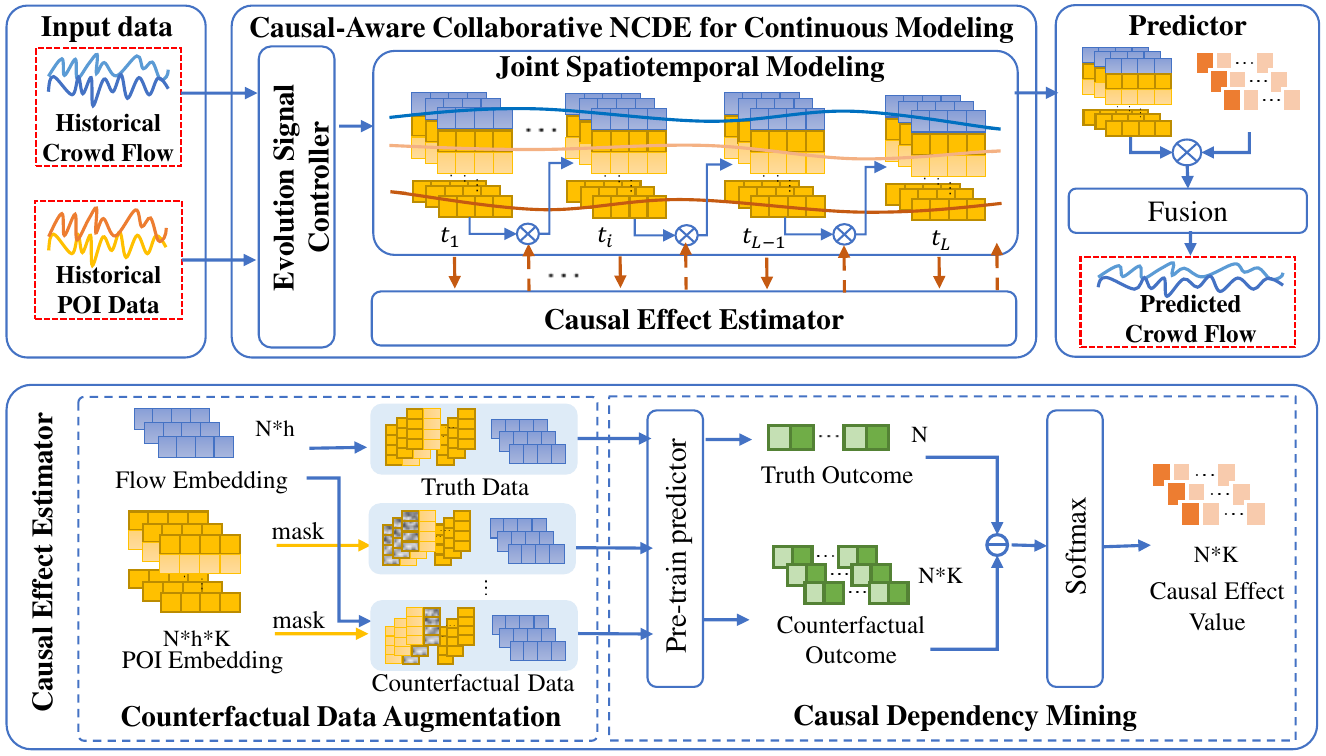}
\caption{Framework overview of C$^3$DE.}
\label{framework}
\vspace{-.5cm}
\end{figure*}

In this section, we introduce the proposed C$^3$DE framework, as shown in Fig. \ref{framework}.
It comprises two main modules. The first is the main pipeline of causal-aware collaborative neural CDE, which models the continuous evolution of collaborative signals while uncovering their potential causal impacts.
The second is the well-designed causal effect estimator, consisting of counterfactual data augmentation and causal dependency mining, designed to explore the causal relationships between collaborative signals.

\subsection{Dual Neural CDE}
We first introduce a naive dual neural CDE as $F(\cdot)$ for modeling both the crowd flow and the POI distribution data simultaneously.
It is formulated as:
\begin{equation}
\begin{cases}
h_x(t^\prime)=h_x(0)+\int_{0}^{t^\prime} f(h_x(t),t;\theta) \mathrm{d} X_t, \quad \quad t^\prime \in [0, T],
\\
h_p(t^{\prime\prime})=h_p(0)+\int_{0}^{t^{\prime\prime}} f(h_p(t),t;\theta) \mathrm{d}P_{t}, \quad \quad t^{\prime\prime} \in [0, M],
\end{cases}
\label{dual cde}
\end{equation}
where $h_x(t^\prime)$ and $h_p(t^{\prime\prime})$ represent the hidden states of crowd flow and POI distribution at $t^\prime$ and $t^{\prime\prime}$ respectively,
which ~\cite{islam2015comparative}.
The control signals $X_t$ and $P_t$ guide the dynamic evolution process of the dual neural CDE, which are derived from the urban crowd flow and the POI distribution, respectively.
Given the crowd flow $X_{t^\prime-T+1:t^\prime} \in \mathbb{R}^{T \times N \times C} $ and POI distribution $P_{t^{\prime\prime}-M+1:t^{\prime\prime}} \in \mathbb{R} ^{M \times N \times K}$, we use the natural cubic spline~\cite{choi2022graph} to create continuous paths which are needed in a neural CDE for control signals:
\begin{equation}
X_t = Spline(X_{t^\prime-T+1:t^\prime}), \quad  P_t = Spline(P_{t^{\prime\prime}-M+1:t^{\prime\prime}}),
\end{equation}
where $Spline(\cdot)$ denotes the natural cubic spline function, which generates continuous, smooth, and twice-differentiable paths for a given input, ensuring accurate and stable gradient computation.

\subsubsection{Exemplification.}
In Eq.(\ref{dual cde}), the function $f(\cdot)$, which deals with the spatio-temporal features of signals, can be applied with any model for processing sequential data.

Without loss of generality, we leverage Gated Recurrent Unit (GRU)~\cite{dey2017gate} as an example to illustrate the derivation of $f(\cdot)$ in Eq.(\ref{dual cde}).
To extend the state update of GRU to the continuous time domain, we introduce the state change $\bigtriangleup h_t$ over the time interval $\bigtriangleup t$, defined as:
\begin{equation}
    \bigtriangleup h_t = h_t - h_{t-\bigtriangleup t}=(1-z_t) \odot (\tilde{h}_t -h_{t-\bigtriangleup t}),
\end{equation}
where $z_t$ and $\tilde{h}_t$ are the intermediate vectors of GRU.
As the time interval $\bigtriangleup t$ tends to $0$, it can be transformed into the differential form of continuous time:
\begin{equation}
    \frac{\mathrm{d} h(t)}{\mathrm{d} t} = (1-z_t) \odot (\tilde{h}_t -h_{t-\bigtriangleup t}).
\end{equation}

Similarly to the GRU model, the state update of any function $f(\cdot)$ for processing sequential data can be extended to the continuous time domain.

\subsection{Causal-aware Collaborative Neural CDE (C$^3$DE)}
Intuitively, there is a tight interaction between long-term evolution of the urban crowd flow and that of the POI distribution. These mutual causalities result in the insufficient modeling in the dual neural CDE which deals with the two collaborative signals respectively.
To explore causal impacts of collaborative signals, we integrate the causal awareness mechanism and propose the causal-aware collaborative neural CDE.
Regarding the accumulation of spurious correlations during continuous evolution, we employ a dynamic correction mechanism in $F(\cdot)$ and rewrite the Eq.(\ref{dual cde}) to alleviate spurious correlations as follows:
\begin{equation}
\begin{cases}

h_x(t^\prime)=h_x(0)+\int_{0}^{t^\prime} f(h_x(t),t;\theta) \mathrm{d} X_t, \quad \quad t^\prime \in [0, T],
\\
h_p(t^{\prime\prime})=h_p(0)+\int_{0}^{t^{\prime\prime}}\mathcal{C} \cdot f(h_p(t),t;\theta) \mathrm{d}P_{t},\ \  \mathcal{C} \in [\mathcal{C}_1,..., \mathcal{C}_L], t^{\prime\prime} \in [0, M].
\end{cases}
\label{c3de}
\end{equation}
where $\mathcal{C}\in \mathbb{R}^{N \times K}$ is the causal impact weight, used to correct the biased POI representations, and $L$ is the number of observation points during the evolution.
We design a counterfactual-based causal effect estimator $g(\cdot)$ to compute $\mathcal{C}$.
Specifically, it takes the hidden states of collaborative signals at observation points during the evolution as input to quantify:
\begin{equation}
    \mathcal{C}_i=g(h_x(t^\prime_i), h_p(t^{\prime\prime}_i)), i\in\{1,...,L\},
\end{equation}
where $\left \{ t^\prime_1, ..., t^\prime_i, ..., t^\prime_L \right \}  \subset [0, T]$ and $\left \{ t^{\prime\prime}_1, ..., t^{\prime\prime}_i, ..., t^{\prime\prime}_L \right \}  \subset [0, M]$ are the evenly spaced observation time points within their intervals. $\mathcal{C}_i$ denotes the causal impact weight of $i$\text{-}th observation point.
Next, we introduce the design of $g(\cdot)$.

\subsubsection{Counterfactual-Based Causal Effect Estimator.}
Inspired by the success of counterfactual data augmentation in natural language processing~\cite{zmigrod2019counterfactual} and dynamic system~\cite{wang2021counterfactual}, we explore the causal impact of POI on crowd flow from a counterfactual perspective, which aims to answer: "How would crowd flow change if a certain POI were changed?".

Specifically, the causal effect estimator consists of two modules: counterfactual data augmentation and causal dependency mining.

\textbf{Counterfactual Data Augmentation}. To answer the above question, we propose a counterfactual data augmentation method based on category-level perturbation, simulating various scenarios of POI changes.
Specifically, given the POI representation $h_p(t^{\prime\prime}_i) \in\mathbb{R}^{N \times K \times H} $ at the $i$\text{-}th observation points, where $H$ denotes the hidden space dimension, the counterfactual data for the $k$-th POI category is constructed as $h_{p^*}^k(t^{\prime\prime}_i)$:
\begin{equation}
    h_{p^*}^k(t^{\prime\prime}_i) = h_p(t^{\prime\prime}_i) \odot M_k,
\end{equation}
where $M_k$ is a perturbation matrix, such as zero-setting, random perturbation, or mean replacement, that controls the category-level perturbation on the $k$\text{-}th POI category. Take the zero-setting perturbation as an example, the perturbation matrix $M_k$ can be defined as:
\begin{equation}
    (M_k)_{n,j,h}=\begin{cases}
0,  & \text{ if } j=k \\
1,  & \text{otherwise},
\end{cases}
\end{equation}
where $n$ is the region index, $j$ is the POI category index, and $h$ is the hidden space dimension index.

We can generate a set of counterfactual POI data by applying perturbations to the $K$ POI categories: $\left \{ h^1_{p^*}(t^{\prime\prime}_i), ..., h^k_{p^*}(t^{\prime\prime}_i),...,h^K_{p^*}(t^{\prime\prime}_i) \right \}$, where each represents the POI representation under a specific POI category's perturbation.
Next, we pair the generated counterfactual POI representation $h^k_{p^*}(t^{\prime\prime}_i)$ with the crowd flow representation $h_x(t^\prime_i)$ to obtain $K$ pairs of counterfactual samples $\mathfrak{D}_{cf}$:
\begin{equation}
    \mathfrak{D}_{cf}=\left \{  \left ( h_x(t^\prime_i),h^k_{p^*}(t^{\prime\prime}_i) \right ) \mid h^k_{p^*}(t^{\prime\prime}_i) \in \left \{ h^1_{p^*}(t^{\prime\prime}_i),...,h^K_{p^*}(t^{\prime\prime}_i) \right \}    \right \},
\end{equation}
the unperturbed POI representation $h_p(t^{\prime\prime}_i)$ and $h_x(t^\prime_i)$ are paired to form the factual sample $\mathfrak{D}_{fact}$:
\begin{equation}
    \mathfrak{D}_{fact}=\left \{ \left ( h_x(t^\prime_i), h_p(t^{\prime\prime}_i) \right )  \right \}.
\end{equation}

\textbf{Causal Dependency Mining}. To evaluate the dynamic impacts of collaborative signals and reveal their causal dependency, we propose a causal dependency mining module based on factual and counterfactual samples.

We first pre-train a predictor $\mathcal{T} (\cdot)$ with the loss function $\ell(\cdot)$. Notably, $\mathcal{T} (\cdot)$  can be any spatiotemporal model, and here we use MTGNN~\cite{wu2020connecting} as the backbone:
\begin{equation}
    \mathcal{T^*} = arg\min_{\mathcal{T}} \ell(\mathcal{T}(X_{t-T+1:t}, P_{t^{\prime}-M+1:t^{\prime}}),X_{t+1:t+S}).
\end{equation}

Next, we sequentially input the counterfactual samples into the predictor $\mathcal{T} (\cdot)$ to obtain the counterfactual output $O_{x, p^*}^k$:
\begin{equation}
    O_{x, p^*}^k=\mathcal{T}(h_x(t^\prime_i), h^k_{p^*}(t^{\prime\prime}_i)) \in \mathbb{R}^N, \quad  \forall (h_x(t^\prime_i), h^k_{p^*}(t^{\prime\prime}_i)) \in \mathfrak{D}_{cf}.
\end{equation}
Meanwhile, we input the factual samples $\left \{ \left ( h_x(t^\prime_i), h_p(t^{\prime\prime}_i) \right )  \right \} \in \mathfrak{D}_{fact}$ into the same $\mathcal{T} (\cdot)$ to obtain the factual output $O_{x, p}^k$, which is considered as an anchor:
\begin{equation}
    O_{x, p}^k=\mathcal{T}(h_x(t^\prime_i), h_p(t^{\prime\prime}_i)) \in \mathbb{R}^N.
\end{equation}

We quantify the causal impact of a specific POI category on crowd flow by the absolute difference between the anchor and counterfactual outputs. 
For the $k$\text{-}th category of POI, the causal effect value is computed as followed:
\begin{equation}
    \mathcal{C}_k(i)= |O_{x, p}^k - O_{x, p^*}^k| \in \mathbb{R}^N.
\end{equation}
A larger causal effect value $\mathcal{C}_k(i)$ indicates significant fluctuations in crowd flow with the changes in $k$\text{-}th POI category, suggesting its key role in flow variation. 

To evaluate the causal impacts of all categories, we apply $Softmax$ function to normalize all causal effect values, obtaining the overall causal effect value at $i$\text{-}th observation point:
\begin{equation}
    \mathcal{C}(i)=Softmax(\mathcal{C}_1(i),...,\mathcal{C}_K(i)) \in \mathbb{R}^{N \times K}.
\end{equation}

By performing the above operation at $L$ observation points, we can capture the dynamic causal impacts of POI on crowd flow.

\subsection{Predictor and Overall Objective}

\subsubsection{Predictor.}
Through the modeling process of C$^3$DE, we obtain the crowd flow representation $h_x(t^\prime)$ and POI distribution representation $h_p(t^{\prime\prime})$, capturing historical evolution and the key causal features for the prediction task.
We fuse their representations to explore the evolution of collaborative signals between POI and crowd flow, resulting in a comprehensive representation $H$ that captures the multidimensional features of crowd flow changes:
\begin{equation}
    H = \sigma (h_x(t^\prime) \cdot W_x+b_x) \odot (h_p(t^{\prime\prime}) \cdot W_p+b_p),
\end{equation}
where $W_x$ and $W_p$ are the learnable weight matrices, $b_x$ and $b_p$ are learnable bias, and $\sigma(\cdot)$ denotes the $sigmod$ function. 

Subsequently, the fused representation $H$ is fed into a multilayer perceptron-based predictor $G(\cdot)$ to predict the next $S$ time steps, as shown below:
\begin{equation}
    \hat{X}_{t+1:t+S} = G(H;\theta _g) \in \mathbb{R} ^{S \times N \times C},
\end{equation}
where $\hat{X}_{t+1:t+S}$ denotes the predicted values.

\subsubsection{Overall Objective.}
Finally, we adopt the Huber loss as the objective function $\ell(\cdot)$. Compared to the traditional squared error loss, it exhibits greater robustness in handling outliers. 
For simplicity, we use $Y$ and $\hat{Y}$ to represent ${X}_{t^\prime+1:t^\prime+S}$ and $\hat{X}_{t^\prime+1:t^\prime+S}$, respectively. 
The learning objective is expressed as:
\begin{equation}
    \ell (Y, \hat{Y})=\begin{cases}
\frac{1}{2}(Y - \hat{Y}),   & |Y - \hat{Y}|\le \delta \\
\delta |Y - \hat{Y}| -\frac{1}{2}\delta ^2,  & \text{otherwise,}
\end{cases}
\end{equation}
where $\delta$ is a hyperparameter that controls the sensitivity to outliers.

\subsection{Complexity Analysis of C$^3$DE}
In the solving process of C$^3$DE, we adopt the adjoint sensitivity method ~\cite{errico1997adjoint} to compute gradients efficiently. Unlike traditional backpropagation, this method solves an auxiliary adjoint differential equation to trace gradients backward in time, requiring only the storage of the final state rather than the entire forward trajectory. This leads to a space complexity of $O(N\cdot d)$, where $N$ denotes the number of nodes and $d$ is the dimension of the hidden state, significantly lower than that of standard backpropagation.
However, this advantage in space comes at the cost of additional computation time. 
Since the adjoint method requires an extra backward integration, the time complexity is approximately $O(2 \cdot N_{fe} \cdot C_f)$, where $N_{fe}$ is the number of times the CDE solver calls the function $f(\cdot)$, and $C_f$ is the time cost of the spatio-temporal modeling function $f(\cdot)$.
Given that our task focuses on long-term crowd flow prediction, where prediction accuracy and stability are prioritized over real-time inference, this trade-off in computation cost is acceptable. 
The advantages in storage space and model performance make our approach both practical and deployable in real-world urban management applications.

%% file: 06.Experiments.tex
\section{Experiments}

\subsection{Experimental setup}
\textbf{Dataset.}
\setcounter{footnote}{0}
We evaluate the proposed framework on three real-world urban crowd flow datasets and their corresponding POI datasets: \textit{NYC-1} and \textit{NYC-2}, collected from NYC OpenData\footnote{https://opendata.cityofnewyork.us/}, and \textit{Beijing}~\cite{zheng2023prediction}.
We summarize the statistics for three datasets in Table \ref{dataset_flow} and Table \ref{dataset_POI}.

\begin{table}[t]
\setlength{\abovecaptionskip}{-2mm} 
\caption{Statistics of urban crowd flow dataset.}
\label{dataset_flow}
\begin{center}
\begin{tabular}{c|c|c|c}
\hline
\textbf{Description}&{\textbf{NYC-1}}&\textbf{NYC-2}&\textbf{Beijing} \\ \hline
time spanning & 2012.06$ \sim $ 2014.05 & 2014.09$ \sim $2016.12 & 2018.07 $\sim$ 2019.10\\
\# of time steps & 16,128 & 17,424 & 10,241\\
\# of records & 2,322,432 & 2,787,840 & 1,894,585\\
\# of nodes & 144 & 160 & 185\\
\hline
\end{tabular}
\end{center}
\vspace{-.8cm}
\end{table}

\begin{table}[t]\label{POI data}
\setlength{\abovecaptionskip}{-2mm} 
\caption{Statistics of POI distribution dataset.}
\label{dataset_POI}
\begin{center}
\begin{tabular}{c|c|c|c}
\hline
\textbf{Description}&{\textbf{NYC-1}}&\textbf{NYC-2}&\textbf{Beijing} \\ \hline
time spanning & 2011.10$ \sim $ 2014.05 & 2014.01$ \sim $2016.12 & 2017.10 $\sim$ 2019.10\\
\# of records & 23,040 & 28,800 & 32,375\\
\# of nodes & 144 & 160 & 185\\
\# of types & 5 & 5 & 7 \\
\hline
\end{tabular}
\end{center}
\vspace{-.5cm}
\end{table}

\noindent 
\textbf{Baselines.}
To evaluate the effectiveness of our C$^3$DE, we compare it with the following baselines:
\begin{itemize}
    \item Traditional methods: 
        \textbf{HA} predicts future values by averaging historical data from the same time period. 
        \textbf{SVR} is a regression method based on support vector machines.

    \item Discrete methods: 
        \textbf{STGCN}~\cite{yu2018spatio} learns spatio-temporal dependencies with a graph convolutional structure.
        \textbf{GWNET}~\cite{wu2019graph} uses an adaptive adjacency matrix to capture hidden spatial dependencies.
        \textbf{STSGCN}~\cite{song2020spatial} captures localized correlations via the synchronous mechanism.
        \textbf{MTGNN}~\cite{wu2020connecting} is a general GNN for modeling multivariate time series.
        \textbf{STWave}~\cite{fang2023spatio} is a decomposition-based framework that decouples flow using wavelet transform.
        
    \item Continuous methods:
        \textbf{STGODE}~\cite{fang2021spatial} extends GNNs with tensor-based ODEs to build deeper networks.
        \textbf{STG-NCDE}~\cite{choi2022graph} designs two NCDEs to model temporal and spatial dependencies.
        \textbf{MTGODE}~\cite{jin2022multivariate} uses NODEs and dynamic graph structure learning to model continuous dynamics.
\end{itemize}

\noindent 
\textbf{Evaluation Metrics.}
We use Mean Absolute Error (MAE), Root Mean Square Error (RMSE) and Mean Absolute Percentage Error (MAPE) to evaluate performance.
Lower values of these metrics indicate better performance.

\noindent 
\textbf{Implementation Details.}
We implemented C$^3$DE in PyTorch using the Adam optimizer with a learning rate $lr=0.001$, weight decay of $5 \times 10^{-4}$, and batch size $B=64$. 
The representation size was fixed to $64$ for all methods. 
We set the historical observation length to $T=14$, $M=4$, and the future prediction length to $S=14$. 
For the Beijing, NYC-1 and NYC-2 datasets, we set the number of observation points to $L=10/8/8$, respectively. 
For counterfactual data augmentation, we applied zero-setting perturbation by default. 
We used an adaptive solver for the Beijing and NYC-2 datasets and the 4th order Runge-Kutta (RK4) solver with a step size of 1.2 for NYC-1.
The codes are available at \href{https://github.com/Sonder-arch/C3DE}{https://github.com/Sonder-arch/C3DE}.

\sethlcolor{gray!30}
\begin{table*}[t]
\captionsetup{skip=2pt}
\setlength{\heavyrulewidth}{1.5pt}

    \caption{Overall performance comparison on three real-world datasets. \hl{Highlighting} denotes the best results and \textbf{bolding} denotes the second-best results.}
\label{Overall result}
\small
\resizebox{\textwidth}{!}{
\begin{threeparttable}
\begin{tabular}{l|c|ccc|ccc|ccc}
\toprule  
\multicolumn{1}{l|}{\multirow{2}*{\textbf{Dataset}}} &
\multicolumn{1}{c|}{\multirow{2}*{\textbf{Method}}} &  \multicolumn{3}{c|}{\textbf{Horizon 7}} & \multicolumn{3}{c|}{\textbf{Horizon 14 }} & \multicolumn{3}{c}{\textbf{Average}} \\ 
\multicolumn{1}{l|}{}  & \multicolumn{1}{l|}{} & MAE & RMSE & MAPE & MAE & RMSE & MAPE & MAE & RMSE & MAPE \\ 
\toprule  
\multicolumn{1}{l|}{\multirow{9}*{\textbf{Beijing}}} &
HA  & 289.40 & 756.21 & 81.4\% & 289.40 & 756.21 & 81.4\% & 289.40 & 756.21 & 81.4\% \\ 
\multicolumn{1}{l|}{} & VAR & 280.41 & 724.90 & 74.2\% & 285.71 & 731.12 & 79.5\% & 283.06 & 728.01 & 76.8\%\\
\cline{2-11} 
\multicolumn{1}{l|}{} & STGCN & 271.55 & 669.84 & 77.3\% & 283.85 & 692.63 & 80.1\% & 277.70 & 681.24 & 78.7\%\\
\multicolumn{1}{l|}{} & GWNET & 170.33 & 414.13 & \textbf{39.0}\% & 208.77 & 503.38 & 47.7\% & 189.55 & 458.76 & 43.4\%\\ 
\multicolumn{1}{l|}{} & STSGCN & 198.82 & 507.28 & 50.7\% & 221.04 & 546.77 & 53.3\% & 209.93 & 527.02 & 52.0\%\\
\multicolumn{1}{l|}{} & MTGNN & 196.19 & 483.25 &  38.9\% & 232.93 & 561.09 &  48.2\% & 214.56 & 522.17 &  43.6\%\\
\multicolumn{1}{l|}{} & STWave & 212.40 & 522.14 &  48.6\% & 242.89 & 588.74 &  54.4\% & 227.65 & 555.44 &  51.5\%\\
\cline{2-11} 
\multicolumn{1}{l|}{} & STGODE & 209.26 & 498.13 &  58.1\% & 234.87 & 560.02 &  59.2\% & 222.07 & 529.08 &  58.7\%\\
\multicolumn{1}{l|}{} & STG-NCDE & \textbf{159.56} & \textbf{404.01} & 39.2\% & \textbf{202.23} & \textbf{491.81} & \textbf{47.0}\% & \textbf{180.89} & \textbf{447.91} & \textbf{43.1}\%\\
\multicolumn{1}{l|}{} & MTGODE & 218.85 & 572.61 &  75.1\% & 222.38 & 594.06 &  76.4\% & 220.61 & 583.33 &  75.7\%\\
\cline{2-11} 
\multicolumn{1}{l|}{} & \rule{0pt}{2.8ex} \textbf{C$^3$DE} & \colorbox{gray!30}{117.56} & \colorbox{gray!30}{245.20} & \colorbox{gray!30}{35.3}\% & \colorbox{gray!30}{124.05} & \colorbox{gray!30}{248.39} & \colorbox{gray!30}{41.9}\% & \colorbox{gray!30}{120.81} & \colorbox{gray!30}{246.80} & \colorbox{gray!30}{38.6}\%\\ 
\bottomrule 
\multicolumn{1}{l|}{\multirow{9}*{\textbf{NYC-1}}} &
HA    & 39.065 & 106.79 & 34.42\% & 39.065 & 106.79 & 34.42\% & 39.065 & 106.79 & 34.42\%\\ 
\multicolumn{1}{l|}{} & VAR & 36.186 & 101.18 & 32.70\% & 37.393 & 104.03 & 33.80\% & 36.789 & 102.60 & 33.25\%\\
\cline{2-11} 
\multicolumn{1}{l|}{} & STGCN & 28.777 & 63.783 & 26.68\% & 30.316 & 77.264 & 26.99\% & 29.546 & 70.524 &  26.84\%\\
\multicolumn{1}{l|}{} & GWNET & 25.060 & 60.958 & 17.40\% & 24.970 & 61.039 & 17.96\% & 25.015 & 60.999 & 17.68\%\\
\multicolumn{1}{l|}{} & STSGCN & 26.143 & 62.917 & 26.19\% & 26.943 & 63.592 & 26.14\% & 26.543 & 63.255 & 26.17\%\\
\multicolumn{1}{l|}{} & MTGNN & 24.165 & 57.958 &  18.94\% & 24.899 & 58.121 &  20.30\% & 24.532 & 58.040 &  19.62\%\\
\multicolumn{1}{l|}{} & STWave & 24.354 & 57.839 &  16.90\% & 24.824 & 58.256 &  17.35\% & 24.589 & 58.047 &  17.13\%\\
\cline{2-11} 
\multicolumn{1}{l|}{} & STGODE & 24.868 & 58.664 &  20.01\% & 25.095 & 58.777 &  20.75\% & 24.982 & 58.721 & 20.38\%\\
\multicolumn{1}{l|}{} & STG-NCDE & 24.693 & 58.289 & 19.53\% & 24.988 & 58.479 & 20.52\% & 24.841 & 58.384 & 20.03\%\\ 
\multicolumn{1}{l|}{} & MTGODE & \textbf{24.141} & \textbf{57.771} &  \textbf{16.37}\% & \textbf{24.758} & \textbf{57.726} &  \textbf{17.06}\% & \textbf{24.449} & \textbf{57.748} &  \textbf{16.72}\%\\
\cline{2-11} 
\multicolumn{1}{l|}{} & \rule{0pt}{2.8ex} \textbf{C$^3$DE} & \colorbox{gray!30}{23.997} & \colorbox{gray!30}{55.598} & \colorbox{gray!30}{14.16}\% & \colorbox{gray!30}{24.050} & \colorbox{gray!30}{56.208} & \colorbox{gray!30}{14.31}\% & \colorbox{gray!30}{24.024} & \colorbox{gray!30}{55.903} & \colorbox{gray!30}{14.24}\%\\ 
\bottomrule 
\multicolumn{1}{l|}{\multirow{9}*{\textbf{NYC-2}}} &
HA    & 24.963 & 63.686 & 24.73\% & 24.963 & 63.686 & 24.73\% & 24.963 & 63.686 & 24.73\%\\ 
\multicolumn{1}{l|}{} & VAR & 23.908 & 61.967 & 16.46\% & 24.232 & 62.761 & 16.22\% & 24.070 & 62.364 & 16.34\%\\
\cline{2-11} 
\multicolumn{1}{l|}{} & STGCN & 19.969 & 34.225 & 11.63\% & 20.619 & 36.281 & 11.64\% & 20.294 & 35.253 & 11.64\%\\
\multicolumn{1}{l|}{} & GWNET & 11.043 & 27.509 & 8.73\% & 11.665 & 29.120 & 8.90\% & 11.354 & 28.314 & 8.82\%\\
\multicolumn{1}{l|}{} & STSGCN & 11.424 & 27.697 & 8.84\% & 11.736 & 29.163 & 8.92\% & 11.580 & 28.430 & 8.88\%\\
\multicolumn{1}{l|}{} & MTGNN & 10.699 & 26.470 &  \textbf{8.18}\% & 11.102 & 28.049 &  8.36\% & 10.901 & 27.260 &  8.27\%\\
\multicolumn{1}{l|}{} & STWave & 10.933 & 26.074 &  8.37\% & 11.676 & 28.056 &  8.45\% & 11.304 & 27.065 &  8.41\%\\
\cline{2-11} 
\multicolumn{1}{l|}{} & STGODE & 12.780 & 28.357 &  11.55\% & 12.780 & 29.633 &  11.43\% & 12.780 & 28.995 &  11.49\%\\
\multicolumn{1}{l|}{} & STG-NCDE & 10.876 & 27.601 & 8.22\% & 11.342 & 28.970 & 8.78\% & 11.109 & 28.286 & 8.50\%\\ 
\multicolumn{1}{l|}{} & MTGODE & \textbf{10.545} & \textbf{26.028} &  \textbf{8.18}\% & \textbf{10.816} & \textbf{28.023} &  \textbf{8.33}\% & x\textbf{10.681} & \textbf{27.026} &  \textbf{8.26}\%\\
\cline{2-11} 
\multicolumn{1}{l|}{} & \rule{0pt}{2.8ex} \textbf{C$^3$DE} & \colorbox{gray!30}{10.159} & \colorbox{gray!30}{25.619} & \colorbox{gray!30}{7.98}\% & \colorbox{gray!30}{10.115} & \colorbox{gray!30}{27.257} & \colorbox{gray!30}{8.07}\% & \colorbox{gray!30}{10.137} & \colorbox{gray!30}{26.438} & \colorbox{gray!30}{8.03}\%\\ 
\bottomrule 
\end{tabular}
\end{threeparttable}
}
\end{table*}

\vspace{-.2cm}
\subsection{Overall Performance}
We evaluate C$^3$DE on three real-world datasets for the task of long-term urban crowd flow prediction, with results in Table \ref{Overall result}.
We observe:
(1) Statistical methods HA and VAR perform the worst, as relying solely on historical data fails to capture complex and dynamic spatiotemporal patterns, leading to significant prediction errors.
(2) MTGODE performs sub-optimally on the NYC-1 and NYC-2 datasets but experiences a sharp performance drop on the Beijing dataset. 
While its continuous-time modeling and dynamic graph structure can effectively capture long-term dependencies in the stable NYC data, it struggles with the complex temporal dynamics of the more volatile Beijing dataset, resulting in instability.
In contrast, STG-NCDE achieves the second-best performance on the Beijing dataset, likely due to its NCDEs-based independent spatio-temporal modeling, which better captures sudden flow changes and intricate temporal dynamics.
(3) Continuous methods do not always outperform discrete methods.
MTGNN consistently surpasses STGODE across all three datasets, likely because while STGODE employs a continuous GNN with residual connections to avoid over-smoothing, it still relies on a fixed graph structure, limiting its ability to capture potential correlations. MTGNN overcomes this limitation with node-adaptive graph convolution.
(4) C$^3$DE consistently outperforms all baselines, especially on the Beijing dataset, demonstrating its superior generalization and stability. It is due to its ability to uncover complex data changes through counterfactual inference. 
When handling highly volatile collaborative signals, it more accurately models their continuous evolution, demonstrating stronger robustness and generalization in complex scenarios.

\vspace{-.4cm}
\subsection{Ablation Study}
\begin{figure*}[t]
\centering  
\subfloat{
\includegraphics[width=0.8\linewidth]{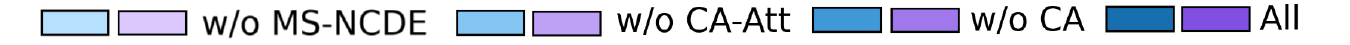}}
\vspace{-.3cm}
\\
\subfloat{
\includegraphics[width=0.33\linewidth]{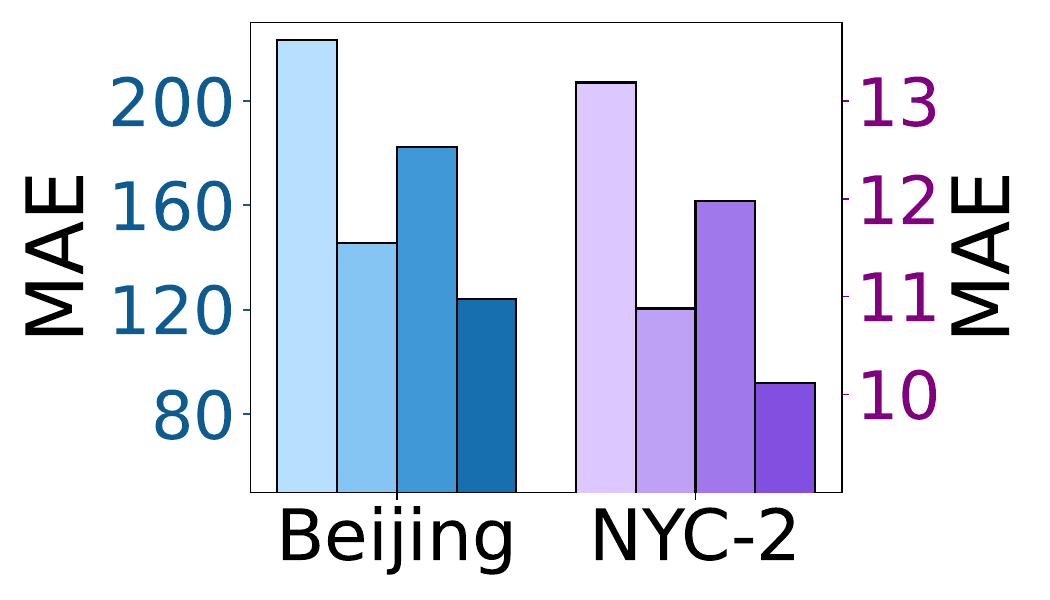}\includegraphics[width=0.33\linewidth]{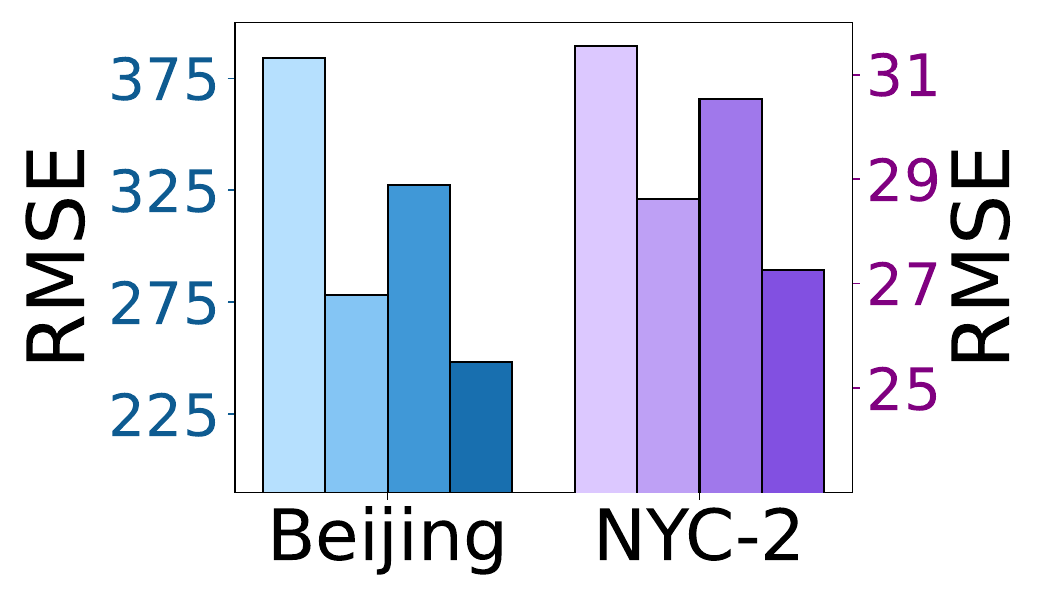}\includegraphics[width=0.33\linewidth]{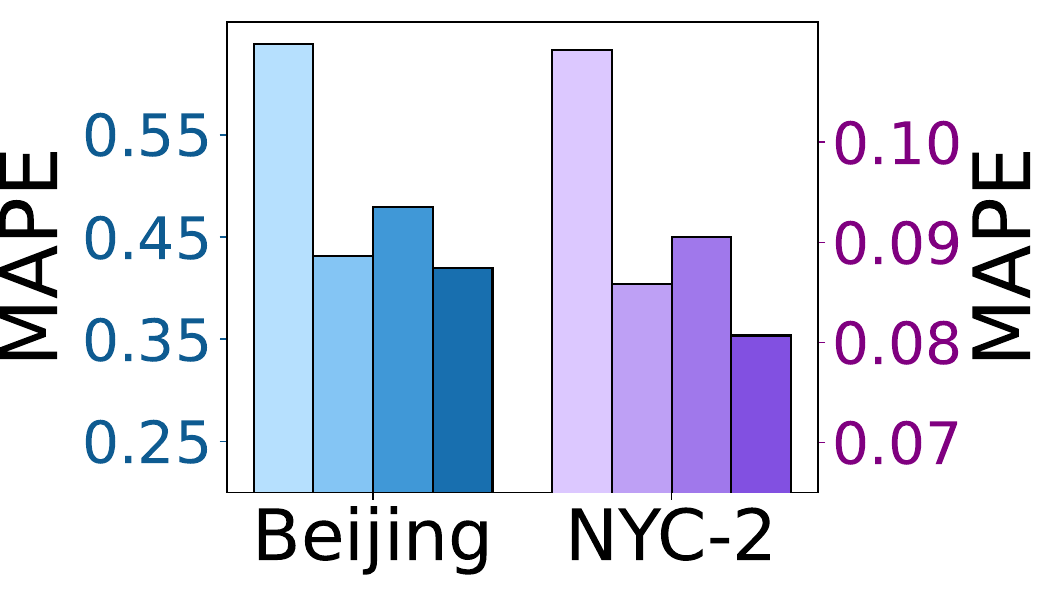}}
\caption{Ablation study on Beijing and NYC-2 datasets.}
\label{ablation}
\vspace{-.5cm}
\end{figure*}

In this section, we further validate the effectiveness of the proposed modules in C$^3$DE, with a particular focus on the continuous modeling and causal mining modules. 
Specifically, we design the following variants, and the experimental results on Beijing and NYC-2 datasets are shown in Fig. \ref{ablation}.
\begin{itemize}
    \item \textit{w/o MS-NCDE}: Remove the NCDE continuous modeling module.
    \item \textit{w/o CA-Att}: Replacing counterfactual-based causal effect values with attention mechanism-based values.
    \item \textit{w/o CA}: Remove the counterfactual-based causal effect estimator module totally, only simply fuse POI and flow final representations.
    \item \textit{All}: It is our complete framework.
\end{itemize}

\noindent
\textbf{Effectiveness of Dynamic Continuous Modeling.}
Experimental results show that \textit{w/o MS-NCDE} performs the worst across the three datasets, highlighting the effectiveness of collaborative NCDE in continuous collaborative signals modeling.
Specifically, collaborative NCDE, formulated as differential equations, can smoothly capture the fine-grained continuous spatio-temporal evolution of collaborative signals, thereby effectively learning potential changes beyond the observation points.

\noindent
\textbf{Effectiveness of Causal Dependency Mining.}
From the results, we can see that the \textit{All} outperforms the \textit{w/o CA} and \textit{w/o CA-Att}, which demonstrates the effectiveness of the counterfactual-based causal effect estimator in capturing and eliminating spurious correlations.
Further, we also find that: first, the \textit{w/o CA} variant performs worst among the three variants, indicating that relying solely on POI distribution for prediction is insufficient. 
While POI distribution can partially reflect flow dynamics, not all POI have a substantive causal relationship with crowd flow. 
Many POIs exhibit only superficial correlations, which introduce spurious relationships and weaken the model's expressiveness, leading to the performance degradation of \textit{w/o CA}.
Second, \textit{w/o CA-Att} models collaborative signals based on attention, dynamically assigning weights to POI distributions to highlight key signals. 
However, it fundamentally relies on data correlations, making it challenging to distinguish true causal relationships. 
Third, \textit{All} employs a counterfactual framework for causal inference, capturing more interpretable causal dependencies and mitigating spurious correlations, leading to superior modeling of collaborative signal evolution.

\begin{table*}[t]
\captionsetup{skip=2pt}
\setlength{\heavyrulewidth}{1.2pt}
\renewcommand{\arraystretch}{1} 
    \caption{The impact of different counterfactual strategies on Beijing dataset.}
\small
\resizebox{\textwidth}{!}{
\begin{threeparttable}
\begin{tabular}{c|ccc|ccc|ccc}
\toprule  
\multicolumn{1}{c|}{\multirow{2}*{\textbf{Method}}} &  \multicolumn{3}{c|}{\textbf{Horizon 7}} & \multicolumn{3}{c|}{\textbf{Horizon 14 }} & \multicolumn{3}{c}{\textbf{Average}} \\ 
\multicolumn{1}{l|}{} & MAE & RMSE & MAPE & MAE & RMSE & MAPE & MAE & RMSE & MAPE \\ 
\toprule  
baseline (\textit{w/o CA})   & 160.66 & 291.87 & 41.8\% & 182.26 & 327.58 & 47.9\% & 171.46 & 309.73 & 44.9\% \\ 
\cline{1-10} 
C$^3$DE-\textit{random} & \textbf{123.12} & \textbf{252.36} & \textbf{35.9\%} & \textbf{136.08} & \textbf{271.18} & \textbf{42.7\%} & \textbf{129.60} & \textbf{261.77 }& \textbf{39.3\%}\\
C$^3$DE-\textit{zero} & \colorbox{gray!30}{117.56} & \colorbox{gray!30}{245.20} & \colorbox{gray!30}{35.3\%} & \colorbox{gray!30}{124.04} & \colorbox{gray!30}{248.39} & \colorbox{gray!30}{41.9\%} & \colorbox{gray!30}{120.81} & \colorbox{gray!30}{246.80} & \colorbox{gray!30}{38.6\%}\\
C$^3$DE-\textit{mean} & 128.53 & 266.86 & 36.2\% & 142.66 & 278.08 & 43.4\% & 135.60 & 272.47 & 39.8\%\\
\bottomrule 
\end{tabular}
\end{threeparttable}
}
\label{counterfactual exp}
\vspace{-3mm}
\end{table*}

\vspace{-.4cm}
\subsection{The Impact of Different Counterfactual Strategies}
In this section, we explore the impact of different counterfactual strategies on prediction performance. 
Specifically, we employ three strategies: "\textit{random}", "\textit{zero}", and "\textit{mean}", against the baseline \textit{w/o CA}, which removes the causal effect estimator module. 
Table \ref{counterfactual exp} presents the results on Beijing dataset, leading to the following findings:
(1) The "\textit{zero}" strategy performs best. As a stringent intervention, it sets the target POI representation to zero, effectively removing its feature information to explore its direct causal impact on crowd flow. 
(2) Unlike "\textit{zero}", the "\textit{random}" strategy introduces random noise to replace the target POI representation.
However, this may introduce uncertainty into the model's causal inference process, leading to suboptimal performance.
(3) The "\textit{mean}" strategy averages all POIs representations except the target and uses this average as its counterfactual representation. 
However, it achieves the lowest performance, possibly because the averaged spatial distribution information blurs the target POI's unique causal effect, making it challenging for the model to capture its true impact.
(4) Notably, all three strategies outperform the baseline "\textit{w/o CA}", 
demonstrating that our framework effectively mines the true causality and thus enhances performance regardless of the intervention strategy.

\vspace{-.5cm}
\subsection{Impacts of Hyper-Parameters}

\begin{figure*}[t]
\centering  
\subfloat{
\includegraphics[width=0.25\linewidth]{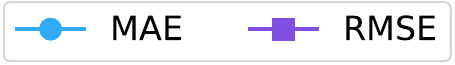}}
\vspace{-.3cm}
\\
\setcounter{subfigure}{0}
\subfloat[Beijing.]{\label{SSBJ}
\includegraphics[width=0.245\linewidth]{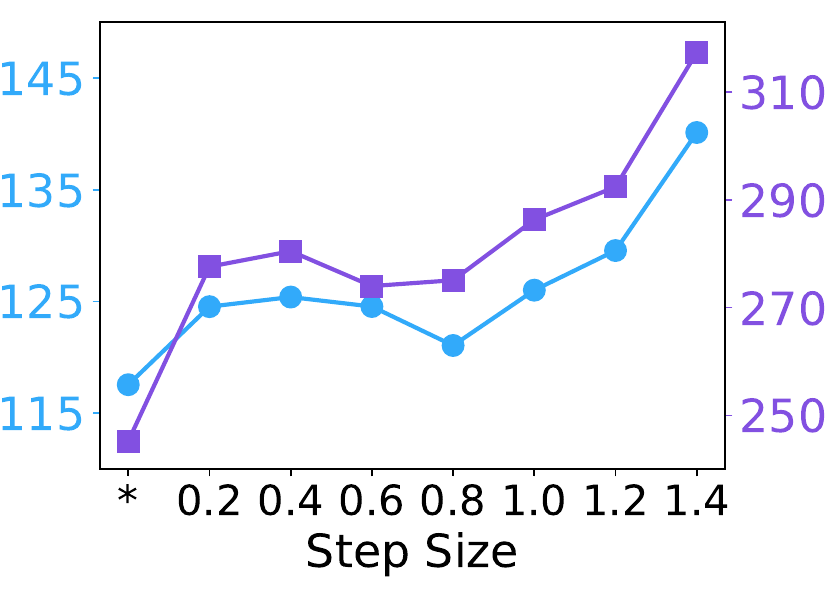}}\subfloat[NYC-1.]{\label{SSNYC}\includegraphics[width=0.245\linewidth]{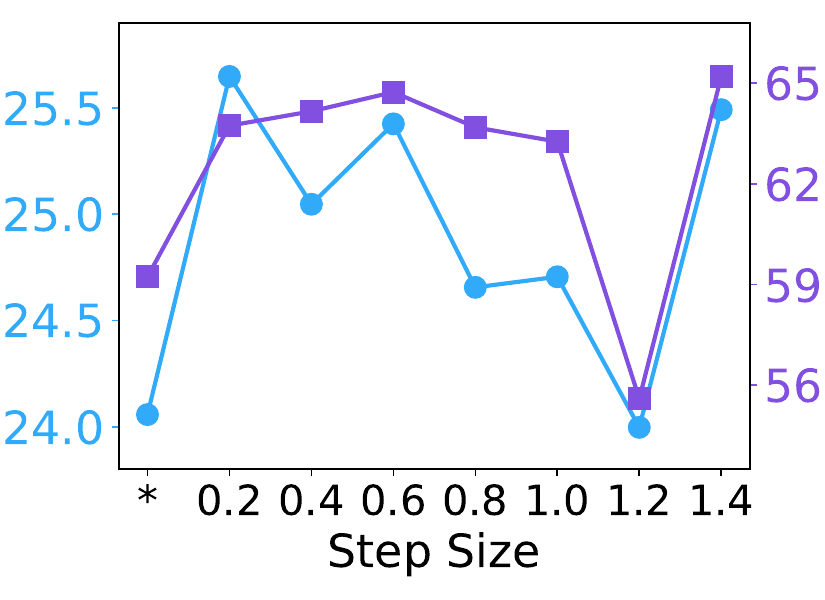}}\subfloat[Beijing.]{\label{LBJ}\includegraphics[width=0.245\linewidth]{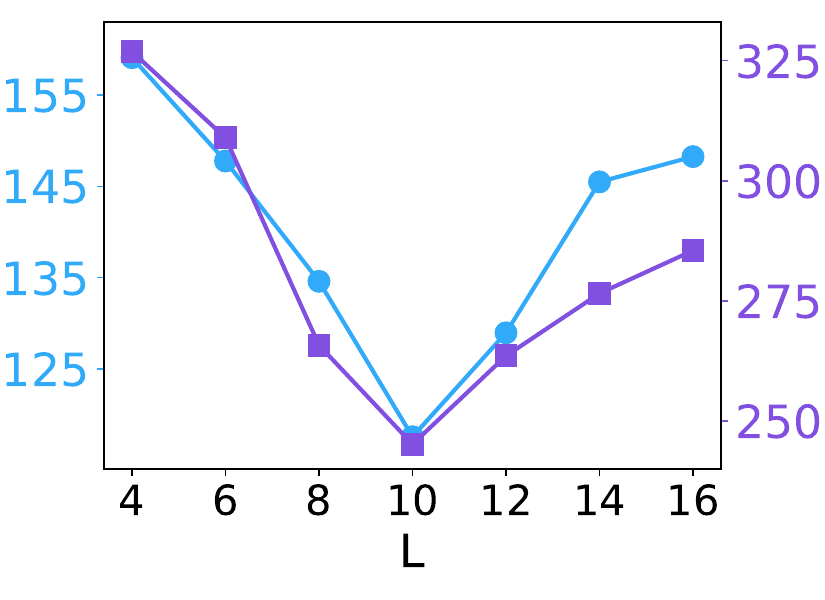}}\subfloat[NYC-1.]{\label{LNYC}\includegraphics[width=0.245\linewidth]{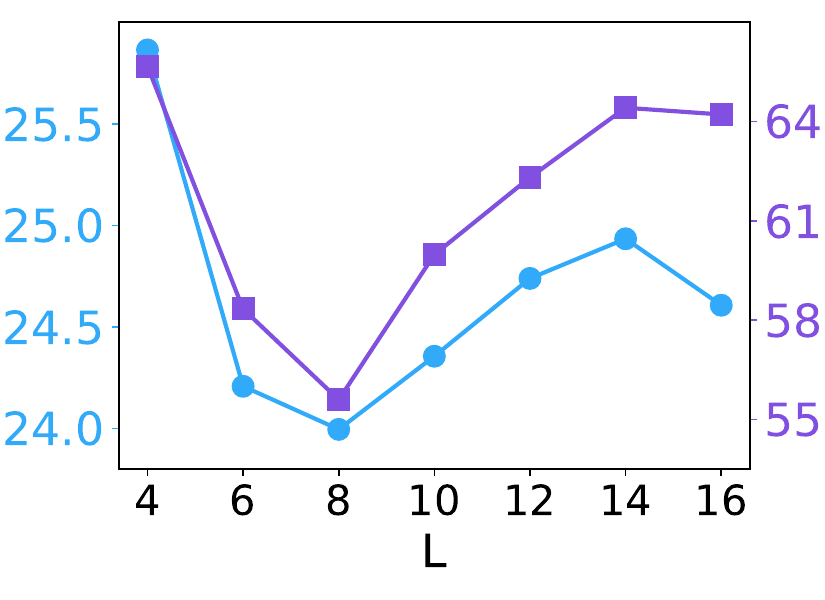}}
\caption{The impacts of \textit{Step Size} and \textit{L} on two datasets.}
\label{hyper}
\vspace{-.5cm}
\end{figure*}

We conduct experiments to validate the impacts of different solvers and the number of observation points $L$ in the dynamic correction mechanism.
First, we choose the adaptive solver and the commonly used fixed-step RK4 solver with different step sizes. Note that, smaller step sizes yield finer data fitting.
As illustrated in Fig. \ref{SSBJ} and Fig. \ref{SSNYC}, the adaptive solver (marked with *) and RK4 solver with a step size of 1.2 perform best on the Beijing and NYC-1 datasets, respectively.  
Performance improves as the step size decreases, as larger step sizes hinder the model to capture precise dynamic changes. 
However, beyond a threshold, further reducing the step size offers no gains, as the variation between each step becomes negligible.
Second, the impacts of varying $L$ from 4 to 16 are shown in Fig. \ref{LBJ} and Fig. \ref{LNYC}. The best results are achieved with $L=10$ for Beijing and $L=8$ for NYC-1.
A small $L$ limits the model's ability to detect and reduce spurious correlations between collaborative signals, while a large $L$ hinders its capacity to capture dynamic signal variations.

\vspace{-.3cm}
\subsection{Visualization of Prediction Results}
\begin{figure*}[t]
\centering  
\subfloat[Node 50 in Beijing.]{
\label{node_50}
\includegraphics[width=0.32\linewidth]{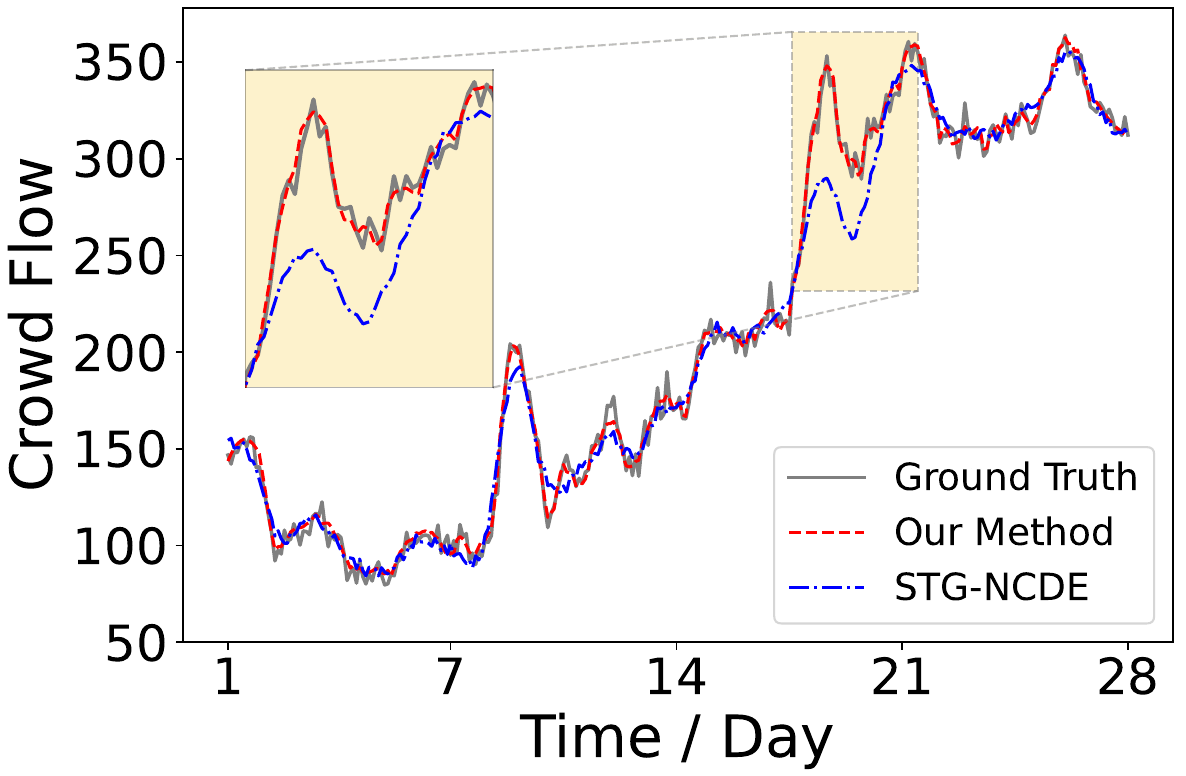}}\subfloat[Node 76 in Beijing.]{
\label{node_76}
\includegraphics[width=0.32\linewidth]{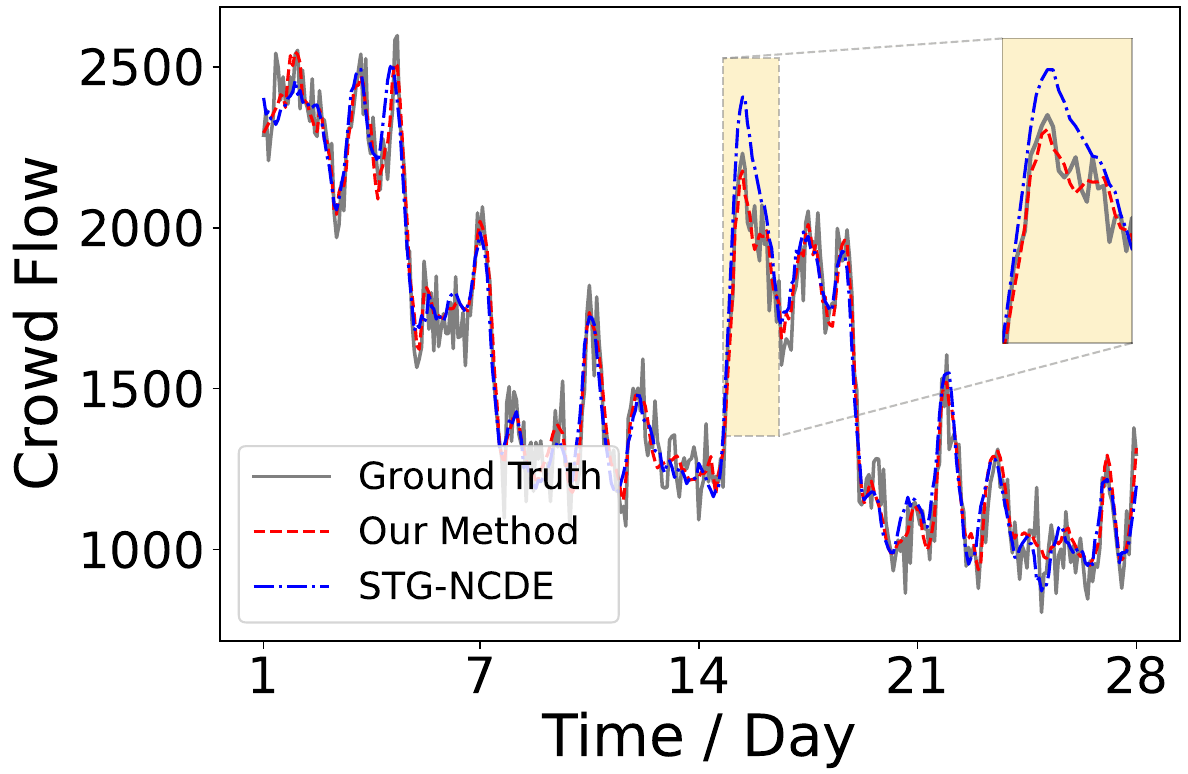}}\subfloat[Node 130 in Beijing.]{
\label{node_130}
\includegraphics[width=0.32\linewidth]{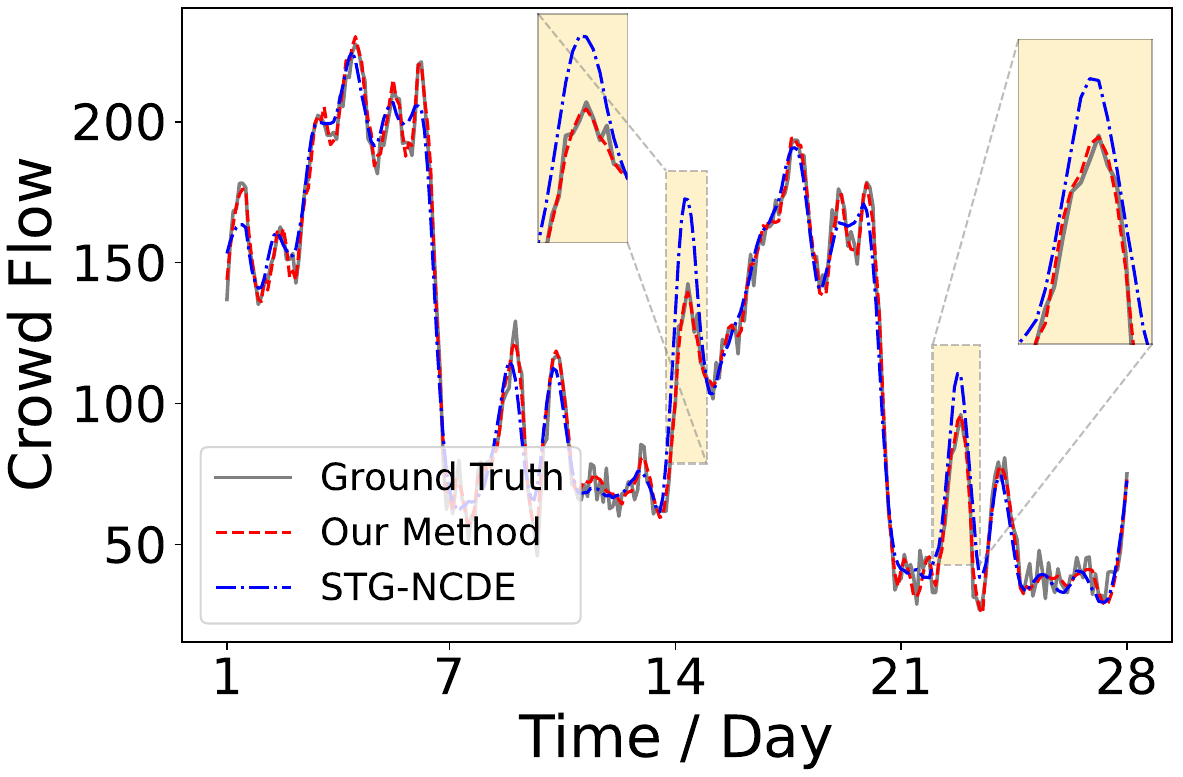}}
\caption{Visualization of prediction results on the Beijing dataset.}
\label{case}
\vspace{-.5cm}
\end{figure*}

We conduct a case to demonstrate the advantages of our method over the continuous modeling baseline, STG-NCDE.
Specifically, C$^3$DE excels at capturing the early-stage changes in fluctuations, which are critical for accurate prediction.
As shown in Fig. \ref{case}, it can not only accurately identify the growth trend at the beginning of the fluctuation(Fig. \ref{node_50}) but also capture the subsequent decline(Fig. \ref{node_76} and Fig. \ref{node_130}), which is due to C$^3$DE's ability to accurately model the relationships between collaborative signals.
These results indicate that C$^3$DE effectively captures the dynamic changes in the system, precisely tracks the early stages of fluctuations, and accurately predicts the future flow variation trends.

\vspace{-.5cm}

%% file: 07.Conclusion.tex
\section{Conclusion}
In this paper, we proposed C$^3$DE, a framework with causal-aware collaborative neural controlled differential equations for long-term urban crowd flow prediction. 
We first introduced the neural CDE with a dual-path architecture to capture the asynchronous dynamic evolution of collaborative signals. 
Next, we designed a counterfactual inference-based causal effect estimator to simulate urban dynamics under different POI distribution scenarios and mine the direct impact of different POIs on crowd flow.
Moreover, we incorporated causal effect values into neural CDE. By introducing a causal effect-based dynamic correction mechanism, C$^3$DE can mitigate the accumulation of spurious correlations among collaborative signals.
Extensive experiments on three real-world datasets demonstrated the significant superiority of C$^3$DE.
Future work will focus on enhancing causal relationship mining efficiency by integrating causal priors based on domain knowledge in urban dynamics.

\vspace{.5cm}
\noindent
\textbf{Acknowledgments.}  
This work was supported by the Natural Science Foundation of China (Grant No.62072235) and the Young Scientists Fund of the Natural Science Foundation of Jiangsu Province, China (Grant No.BK20241402).
We would also like to thank the anonymous reviewers for giving us useful and constructive comments. Additionally, we are grateful to the community and everyone who made their datasets and source codes publicly accessible. These datasets and source codes are valuable and have greatly facilitated this research.